\documentclass[journal,fontsize=10]{IEEEtran}
\ifCLASSINFOpdf
\else
\fi

\usepackage{amsmath,amssymb,amsfonts}
\usepackage{subcaption}
\usepackage{caption}
\usepackage{multirow}
\usepackage{epsfig}
\usepackage{float}
\usepackage{epsfig}
\usepackage{bbm}

\makeatletter
\def\ps@IEEEtitlepagestyle{
  \def\@oddfoot{\mycopyrightnotice}
  \def\@evenfoot{}
}
\def\mycopyrightnotice{
  {\footnotesize
  \begin{minipage}{\textwidth}
  This work has been submitted to the IEEE for possible publication. Copyright may be transferred without notice, after which this version may no longer be accessible.
  \end{minipage}
  }
}

\makeatletter

\begin{document}
%
\title{Towards A Conceptually Simple Defensive Approach for Few-shot classifiers Against Adversarial Support Samples}
%
%
%
\author{Yi~Xiang~Marcus~Tan, 
        Penny~Chong,
        Jiamei~Sun,
        Ngai-Man~Cheung, 
        Yuval~Elovici
        and~Alexander~Binder
\thanks{This research is supported by both ST Engineering Electronics and National Research Foundation, Singapore, under its Corporate Laboratory @ University Scheme (Programme Title: STEE Infosec-SUTD Corporate Laboratory). (Corresponding author: Yi Xiang Marcus Tan, e-mail:marcustanyx16@gmail.com).}
\thanks{Yi Xiang Marcus Tan, Penny Chong, Jiamei Sun was with the Information Systems Technology and Design Pillar, Singapore University of Technology and Design, Singapore 487372, Singapore.}%
\thanks{Ngai-Man Cheung is with the Information Systems Technology and Design Pillar, Singapore University of Technology and Design, Singapore 487372, Singapore.}%
\thanks{Yuval Elovici is with the Department of Software and Information Systems Engineerng, Ben-Gurion University of the Negev, 653 Beer-Sheva, Israel.}
\thanks{Alexander Binder is with the Department of Informatics, Digital Signal Processing and Image Analysis group of the University of Oslo, 0373 Oslo, Norway.}
}

\maketitle

\begin{abstract}

Few-shot classifiers have been shown to exhibit promising results in use cases where user-provided labels are scarce. These models are able to learn to predict novel classes simply by training on a non-overlapping set of classes. This can be largely attributed to the differences in their mechanisms as compared to conventional deep networks. However, this also offers new opportunities for novel attackers to induce integrity attacks against such models, which are not present in other machine learning setups. In this work, we aim to close this gap by studying a conceptually simple approach to defend few-shot classifiers against adversarial attacks. More specifically, we propose a simple attack-agnostic detection method, using the concept of self-similarity and filtering, to flag out adversarial support sets which destroy the understanding of a victim classifier for a certain class. Our extended evaluation on the miniImagenet (MI) and CUB datasets exhibit good attack detection performance, across three different few-shot classifiers and across different attack strengths, beating baselines. Our observed results allow our approach to establishing itself as a strong detection method for support set poisoning attacks. We also show that our approach constitutes a generalizable concept, as it can be paired with other filtering functions. Finally, we provide an analysis of our results when we vary two components found in our detection approach.
\end{abstract}


%
\IEEEpeerreviewmaketitle

\section{Introduction}

%
%
%
%

 




\IEEEPARstart{A}{n} open topic in machine learning is the transferability of a trained model to a new set of prediction categories without retraining efforts, in particular when some classes have very few samples.
Few-shot learning algorithms have been proposed to address this. 
Prediction and training in few-shot approaches are based on the concept of an episode. Each episode (task) comprises several labelled training samples per class (i.e. 1 or 5), denoted as the support set, and query samples for episodic testing called the query set.
Unlike conventional machine learning setups, the prediction in few-shot models is relative to the support set classes of an episode \cite{snell2017prototypical,MAML:finn2017model,SNAIL:mishra2018a, Reptile:nichol2018reptile, MTL:sun2019meta}.
The label categories drawn in each episode varies and training is performed by on these randomised sets of classes. This allows for the iteration over varying prediction tasks when learning model parameters.
Effectively, this learns a class-agnostic similarity metric which allows for generalisation to novel categories \cite{RN:sung2018learning,DeepEMD:zhang2020deepemd,CAN:hou2019cross}.

Unfortunately, the adversarial susceptibility of models under the few-shot classification setting remains relatively unexplored, albeit gaining traction \cite{xu2020meta,goldblum2020adversarially}. This is compared to models under the standard classification setting, where such a phenomenon had been widely explored \cite{szegedy2013intriguing,Papernot2016,carlini2017towards,madry2017towards,tanay2016boundary,Goodfellow2014}.
The relative nature of predictions in few-shot setups allows going beyond crafting adversarial test samples.

The attacker could craft adversarial perturbations for all $n$-shot support samples of the attacked class and insert them into the deployment phase of the model. The goal is to misclassify test samples of the attacked class regardless of the samples drawn in the other classes. In this work, we consider the impact on the few-shot accuracy of the attacked class, in the presence of adversarial perturbations, even when different samples were drawn for the non-attacked classes. This is a highly realistic scenario as the victim could unknowingly draw such adversarial support sets during the evaluation phase once they were inserted by the attacker. The use of adversarial samples to attack other settings than the one trained for are known as transferability attacks.

\IEEEpubidadjcol
In order to mitigate the adverse effects of adversarial attacks, several methods were proposed in the past. Such approaches aim to do so through detection \cite{xu2017feature,Tian2018,cintasdetecting} or through model robustness \cite{zhang2019theoretically,jeddi2020learn2perturb,folz2020adversarial,zhang2020interpreting}. Though these methods work well for neural networks under the conventional classification setting, 
they will fail on 
few-shot classifiers due to the limited data issue that few-shot learners excel at. 
Furthermore, these defences were not trained to transfer its pre-existing knowledge towards a novel distribution of class samples, contrary to few-shot classifiers.
With the aforementioned drawbacks in mind, we propose a conceptually simple method for performing attack-agnostic detection of adversarial support samples in this setting. 
We exploit the concept of support and query sets of few-shot classifiers to measure the similarity of samples within a support set after filtering, for example by autoencoders. 
We perform this by
randomly splitting the original support set randomly into auxiliary support and query sets, followed by filtering the auxiliary support and predicting the query.
If the samples are not self-similar, then we will flag the support set as adversarial.
To this end, we describe the contributions of this work as follows:
\begin{enumerate}
    \item We propose a simple yet novel attack-agnostic detection mechanism against adversarial support sets in the domain of few-shot classification. This is based on self-similarity under randomised splitting of the support set and filtering, and is the first, to the best of our knowledge, for the detection of adversarial support sets in few-shot classifiers. In particular, we show that few-shot learning can be equipped with strong detection approaches for support set poisoning attacks. 
    \item We analyse the effectiveness of such a detection approach, when using various filtering functions of differing mechanisms. To provide a form of comparison, we adopted two simple baselines with one being an unsupervised approach while the other being a supervised method.
    \item We investigate the effects of a unique white-box adversary against few-shot frameworks, through the lens of transferability attacks. Rather than crafting adversarial query samples similar to standard machine learning setups, we optimise adversarial supports sets, in a setting where all non-target classes are varying. Various attack strengths were explored to analyse the trend in our detection performance.
    \item We provide further analysis on the detection performance of our algorithm when using different different filtering functions  and also different formulation variants of the aforementioned self-similarity quantity. Our analysis establishes our proposed approach of self-similarity and filtering as a generalisable concept.
\end{enumerate}

This work extends our prior work in \cite{tan2021detection}, where we introduce here an additional model in our experiments, namely the Prototypical Network (PN) \cite{PROTO:snell2017prototypical}, to improve the generalisability of our approach to other variants of few-shot models. Furthermore, we explored additional filtering functions and baselines, to analyse variations in detection performance of our approach and to provide a simple benchmark. 
More specifically, we describe our newly explored baselines, namely the Out-of-Distribution Image Detection (ODIN) and Isolation Forest (IF) approaches in Sections \ref{odin} and \ref{isolate}, while introducing the Total Variation Minimisation (TVM) and Bit Reduction (BitR) in Sections \ref{tvm} and \ref{bitr} as additional filters. We have also increased the scope of the attack strengths we considered, to show the trends of our detection performance at various scenarios. Comparing to our previous work in \cite{tan2021detection}, we have vastly expanded our experiment settings. 
In Section \ref{transferatt}, we show an extension of our transferability attack analysis by introducing varying degrees of attack strength, for the two attack variants that were explored in this work. Furthermore, we explain our motivation for using self-similarity here, which was missing in our prior work \cite{tan2021detection}. In Section \ref{detectadvsupports}, we further extend our analysis to show the trends of our detection performance. We evaluated across the various attack strengths and attack approaches, filtering functions and baselines, the few-shot models, and datasets.

\section{Related Works}
\label{fewshotrelated}

\subsection{Few-shot classification}
With the aim of mitigating the high demand for labelled data for deep neural networks (DNNs), few-shot classification recognises novel categories with only a few labelled samples per class for training. Notable methods include metric-based classifiers \cite{RN:sung2018learning,CAN:hou2019cross,TAFSSL:10.1007/978-3-030-58571-6_31,DeepEMD:zhang2020deepemd} and optimisation-based classifiers \cite{MAML:finn2017model,MTL:sun2019meta}.
Optimisation-based classifiers learn the initialisation parameters that can quickly generalise to novel categories or train a meta-optimiser that adaptively updates the model parameters for novel classes.  
Metric-based classifiers learn a distance metric that compares the representations of images and generates similarity scores for classification, which made significant progress recently.

\subsection{Poisoning of Support Sets}
There is limited literature examining the poisoning of support sets in meta-learning.
\cite{xu2020meta} proposed an attack routine, Meta-Attack, extending from the highly explored Projected Gradient Descent (PGD) attack \cite{madry2017towards}. They assumed a scenario where the attacker is unable to obtain feedback from the classification of the query set. Hence, the authors used the empirical loss on the support set to generate adversarial support samples which hope to induce misclassification behaviours to unseen query sets.

\subsection{Autoencoder-based and Feature Preserving-based Defences}
There are two recent prior works which utilise autoencoders as to formulate their defence approach. \cite{cintasdetecting} performs the detection of such attacks using Non-parametric Scan Statistics (NPSS), based on hidden node activations from an autoencoder. This NPSS score measures how anomalous a subset of a node activation is, given an input sample.
The authors compute such activations from both clean and adversarial images, compares them and compute the NPSS score. 
\cite{folz2020adversarial} proposed using an autoencoder to reconstruct input samples such that only the necessary signals remain for classification. Their training method is a two-step process, first performing unsupervised training for reconstruction throughout the autoencoder, and the second, training only the decoder based on the classification loss of the input with respect to the ground truth. However, under the few-shot setting, fine-tuning based on the classification loss should be avoided because we would require large enough samples from each class for the fine-tuning step. 
\cite{zhang2020interpreting} attempts to stabilise sensitive neurons which might be more prone to the effects of adversarial perturbations, by enforcing similar behaviours of the sensitive neurons between clean and adversarial inputs through Sensitive Neurons Stabilising (SNS). As such, SNS tries to preserve the features between the reconstructed image and the original image for adversarial robustness by training their defended model, regularised on a feature preserving loss term between clean and adversarial features. The method in \cite{zhang2020interpreting} requires adversarial samples during the training process which potentially makes defending against unseen attacks challenging, since they were unseen during training. In light of this, we proposed a detection approach which does not make use of any adversarial samples. Though we employed the concept of feature preserving as one of our various filtering functions (introduced later in Section \ref{fewshotmethod}), our approach is still different from \cite{zhang2020interpreting} as it does not suffer from this limitation.
Hence, in our work, we adopted an approach which does not require labelled data to train our autoencoder for reconstruction, which we will elaborate further in Section~\ref{fewshotmethod}.

\section{Background}
\label{background}

\subsection{Few-shot classifiers Used}

A majority of the 
few-shot classifiers are trained with \emph{episodes} sampled from the training set. Each episode consists of a support set $S=\{x_s, y_s\}_{s=1}^{K*N}$ with $N$ labelled samples per $K$ classes, and a query set $Q=\{x_q\}_{q=1}^{N_q}$ with $N_q$ unlabelled samples from the same $K$ classes to be classified, denoted as a $K$-way $N$-shot task. The metric-based classifiers learn a distance metric that compares the features of support samples $x_s$ and query sample $x_q$ for classification. During inference, the episodes are sampled from the test set that has no overlapping categories with the training set.

In this work, we explored three known metric-based few-shot classifiers, namely the RelationNet (RN) \cite{RN:sung2018learning}, the Prototypical Network (PN) \cite{snell2017prototypical}, and a state-of-the-art model, the Cross-Attention Network (CAN) \cite{CAN:hou2019cross}.
The support and query samples are first encoded by a backbone CNN to get the image features \{$f^c_s \ | \ c=1,  \dots , K$\} and $f_q$, respectively. 
The feature vectors $f_s^c$ and $f_q \in \mathbbm{R}^{d_f, h_f, w_f}$, 
where $d_f$, $h_f$, and $w_f$ are the channel dimension, height, and width of the image features.
If $N>1$, $f^c_s$ will be the averaged feature of the support samples from class $c$.
To measure the similarity between $f_s^c$ and $f_q$, the RN model concatenates $f_q$ and $f_s^c$ along the channel dimension pairwise and uses a \emph{relation module} to calculate the similarities. The CAN model adopts a \emph{cross-attention} module that generates attention weights for every $\{f_s^c, f_q\}$ pair. The PN model computes the mean of the support samples after extracting their features, for each way of the episode. The attended image features are further classified with cosine similarity in the spirit of dense classification \cite{DENSECLASSIFICATION:lifchitz2019dense}.

\subsection{Threat Model}

We assume that the attacker wants to invalidate the few-shot classifier's notion of a targeted class, $t$, unlike conventional machine learning frameworks where one is optimising single test samples to be misclassified. The attacker wants to find an adversarially perturbed set of support images, such that misclassification of most query samples from class $t$ occurs, regardless of the class labels of the other samples. He then replaces the defender support set for class $t$ with the adversarial support. We assume that the attacker has white-box access to the few-shot model (i.e. weights, architecture, support set).
The adversarial support set would classify itself as self-similar, that is, they classify among each other as being within the same class, visually appear as class $t$, but classify true query images of class $t$ as belonging to another class.

We now clarify our definition of $x$ used in our attacks. The attacks are applied on a fixed support set candidate $(x_1^{t}, \ldots, x_{n_{shot}}^t)$ for the target class. In every iteration of the gradient-based optimisation, we sample all classes randomly except for the target class.
Specifically, we sample the support sets $S^{-t}$ and query sets $Q^{-t}$ of all the other classes randomly, and we randomly sample the query samples of the target $Q^t$, illustrated in the equations below. They are redrawn in every iteration of the optimisation following a uniform distribution. 
\begin{equation}
\begin{split}
 \mathcal{C}^{-t} &\sim Uniform(\mathcal{C}~\backslash~\{t\}) \\
     S^{-t}, Q^{-t} &\sim Uniform(x | c \in \mathcal{C}^{-t}), \ Q^t \sim Uniform(x | c = t)\\
         x &=(x_1^{t}, \ldots, x_{n_{shot}}^t), \\
    h(x)&=   h([x_1^{t}, \ldots, x_{n_{shot}}^t, S^{-t}], [Q^t, Q^{-t}]), 
    \end{split}
    \label{threatmodel}
\end{equation}
where $\mathcal{C}$ is the set of all classes, and $\mathcal{C}^{-t}$ the random set of classes used in the episode together without class $t$ (the cardinality of $\mathcal{C}^{-t}$ is $K-1$ given a $K$-way problem). Here, $h$ is a few-shot classifier that takes in a support set and a query set to return class prediction scores for the query samples. 
The last line in \eqref{threatmodel} indicates that the few-shot classifier $h$ takes in a support set made up of $x$ and $S^{-t}$ and a query set made up of $Q^t$ and $Q^{-t}$, which is a simplification to the expression, to relate to \eqref{chp2:eqnpgd} and \eqref{chp2:eqncw}.
The adversarial perturbations $\delta$ and the underlying gradients are computed only for each of the support samples $x$ of the target class.

\subsection{Attack Algorithms Used}

\subsubsection{Projected Gradient Descent (PGD)}
\label{chp2:pgd}

The Projected Gradient Descent (PGD) attack \cite{madry2017towards} computes the sign of the gradient of the loss function with respect to the input data as adversarial perturbations. For an adversarial candidate $x_i$ at the $i^{th}$ iteration:
\begin{gather}
    x_0 = x_{original} + Uniform([-\epsilon, \epsilon]^d), \\
    x_i = Clip_{x, \epsilon}\{x_{i-1} + \eta \, sign(\nabla_{x} L(h(x_{i-1}), c))\},
    \label{chp2:eqnpgd}
\end{gather}
where $h(.)$ is the prediction logits for classifier $h$ of some adversarial candidate $x_t$, $L$ is the loss used during training (i.e. cross-entropy with softmax for image classification), $\nabla_x L$ represents the gradients of the loss calculated with respect to $x_t$, $\eta$ is the step size and $\epsilon$ is the adversarial strength which limits the adversarial candidate $x_i$ within an $\epsilon$ bounded $\ell_\infty$ ball. Before the iterative perturbation step, the attacker initialises the starting adversarial candidate as a point uniformly sampled around $x_{original}$, bounded by $\epsilon$. This maximises the loss of the adversarial candidate with respect to the source class $c$, which aims to cause a misclassification.


\subsubsection{Carlini \& Wagner $\mathcal{L}_2$ (CW-$\mathcal{L}_2$)}
\label{chp2:cw}
The Carlini \& Wagner $\mathcal{L}_2$ (CW-$\mathcal{L}_2$) attack \cite{carlini2017towards} finds the smallest $\delta$ that successfully fools a target model using the Adam optimiser. Among the various white-box attacks, this approach is known to be highly effective in obtaining successful adversarial samples while achieving a low adversarial perturbation magnitude. However, contrary to the previous white-box attacks, the CW-$\mathcal{L}_2$ is a specifically targeted attack as it involves computing the difference in logits between the targeted prediction and the current highest scoring logit (which is not $t$). Their attack solves the following objective function:
\begin{equation}
    \begin{split}
    &\min_\delta ||\delta||_2 + const\cdot L(x+\delta, \kappa),\\
    s.t.~L(x', \kappa) &= \max(-\kappa, \max_{i \neq t}(h(x')_{i}) - h(x')_t).
    \end{split}
    \label{chp2:eqncw}
\end{equation}
The first term penalises $\delta$ from being too large by minimising the $L_2$ norm of $\delta$ while the second term enforces misprediction. The value $const$ is a weighting factor that controls the trade-off between finding a low $\delta$ and having a successful misprediction. $h(\cdot)_i$ refers to the logits of prediction index $i$ and $t$ refers to the target prediction. $\kappa$ is the confidence value that influences the logits score differences between the target prediction $t$ and the next best prediction $i$.

\subsection{Baseline Algorithms Used}
In our work, we used two algorithms designed to detect out-of-distribution samples, namely Out-of-Distribution Image Detection (ODIN) and Isolation Forest (IF), which can be exploited to detect adversarial samples.. 

\subsubsection{Out-of-Distribution Image Detection (ODIN)}
\label{odin}
The ODIN approach \cite{liang2018enhancing} is a two-step process, first involving a preprocessing step and with the second performing detection based on the maximum scaled softmax probability score from the classifier. Their approach claims that introducing a small perturbation and using temperature-scaled softmax probability scores can make in- and out-of-distribution samples more distinguishable. The scaled softmax probabilities are based on some temperature parameter, $T$, such that:
\begin{gather}
    \mathrm{softmax}_i(x,T) = \frac{\mathrm{exp}(h_i(x)/T)}{\sum_{j=0}^{C-1} \mathrm{exp}(h_j(x)/T)},
    \label{odineqn}
\end{gather}
for $i \in [0,C)$ where $C$ is the total number of classes. $h$ is our classifier function that returns the prediction scores of the input sample. For the preprocessing of the inputs, they adopted a similar approach introduced by \cite{Goodfellow2014} to perform a small perturbation to increase the scaled softmax score. However, they performed standard gradient descent to minimise the loss incurred (i.e. cross-entropy with softmax), instead of the standard gradient ascent, commonly used in adversarial attacks. More specifically, the preprocessed input, $\Tilde{x}$ is computed as follows:
\begin{gather}
    \Tilde{x} = x - \epsilon_{ODIN} \cdot sign(\nabla_x -\mathrm{log}(\mathrm{softmax}_{\Tilde{y}}(x, T))),
\end{gather}
where $\Tilde{y}$ is the hard label prediction of the classifier $h$.
After which, $\Tilde{x}$ is passed as inputs to $h$, where the maximum scaled softmax probability score is computed. The input is flagged as an outlier if this maximum if lower than a certain threshold value, calibrated based on some desired True Positive Rate. The authors noted that good detection rates occur at temperature values above 100 (i.e. $T > 100$), though any value of T above that range does not significantly improve the detection further. 

\subsubsection{Isolation Forest (IF)}
\label{isolate}
The IF, analogous to Random Forest, uses a collection of isolation trees (binary trees) that partitions the data based on randomly selected features of randomly selected values (between the minimum and maximum value of the selected feature) \cite{liu2012isolation}. The algorithm works under the observation that clean data performs deeper traversals from the root of the isolation trees to the leaves, in contrast to anomalous data points, since they are scarce (i.e. lying further away in the feature space as compared to clean data). Consequently, anomalous data lies closer to the root nodes of the isolation trees. The anomalous score of a sample, $x$, being fed to an isolation tree is given as such:
\begin{gather}
    s(x,M) = 2^{-\mathbb{E}(\mathrm{height}(x))/c(M)}
\end{gather}
where $\mathrm{height}(x)$ is the length of the path from the root to the leaf for some input $x$, $c(M)$ is the average number of unsuccessful searches in a binary search tree given $M$ number of nodes. A score closer to 1 exhibit strong abnormality while scores much smaller than 0.5 indicates otherwise.

\section{Detection Methodology}
\label{fewshotmethod}

Our detection-based defence is based on three components: a sampling of auxiliary query and support sets, filtering the auxiliary support sets and measuring an adversarial score with respect to the unfiltered auxiliary query set. 
Our motivation is derived from a high self similarity phenomenon under attacks (high classification accuracy of adversarial samples), which was absent under normal cases (see Table \ref{tab:selfsim}). 
As such, should filtering be performed on the auxiliary support set, we postulate that the auxiliary query set will be less self similar to its auxiliary supports. In order to showcase the generalisability of using self similarity and filtering to detect adversarial samples, we adopted several filtering functions with highly different mechanisms. We denote a statistic either averaged over all possible splits or for a randomly drawn split of a support set into auxiliary sets with filtering of the auxiliary supports as self-similarity. 



\subsection{Auxiliary Sets}
\label{auxset}
Support and query sets in few-shot classifiers can be chosen freely, which implies that any specific sample can be chosen to be part of the support or the query.
Assuming we have a support set for class $c$, we randomly split it into its auxiliary support and query sets, where $S^c$ might be clean or adversarial:
\begin{gather}
    \nonumber S^{c}_{aux} \cup Q^{c}_{aux} = S^c~\text{and}~S^{c}_{aux} \cap Q^{c}_{aux} = \emptyset,\\
    \text{such that}~|S^{c}_{aux}| = n_{shot}-1~\text{and}~|Q^{c}_{aux}| = 1.
\end{gather}
The few-shot learner is now faced with a randomly drawn ($n-1$)-shot problem, evaluating on one query sample per way, with the option to average the $n$ possible splits.






\subsection{Detection of Adversarial Support Sets}
\label{detectsec}

\begin{figure}[htb]
    \centering
    \includegraphics[trim={0 0 0 0},clip,width = 0.48\textwidth]{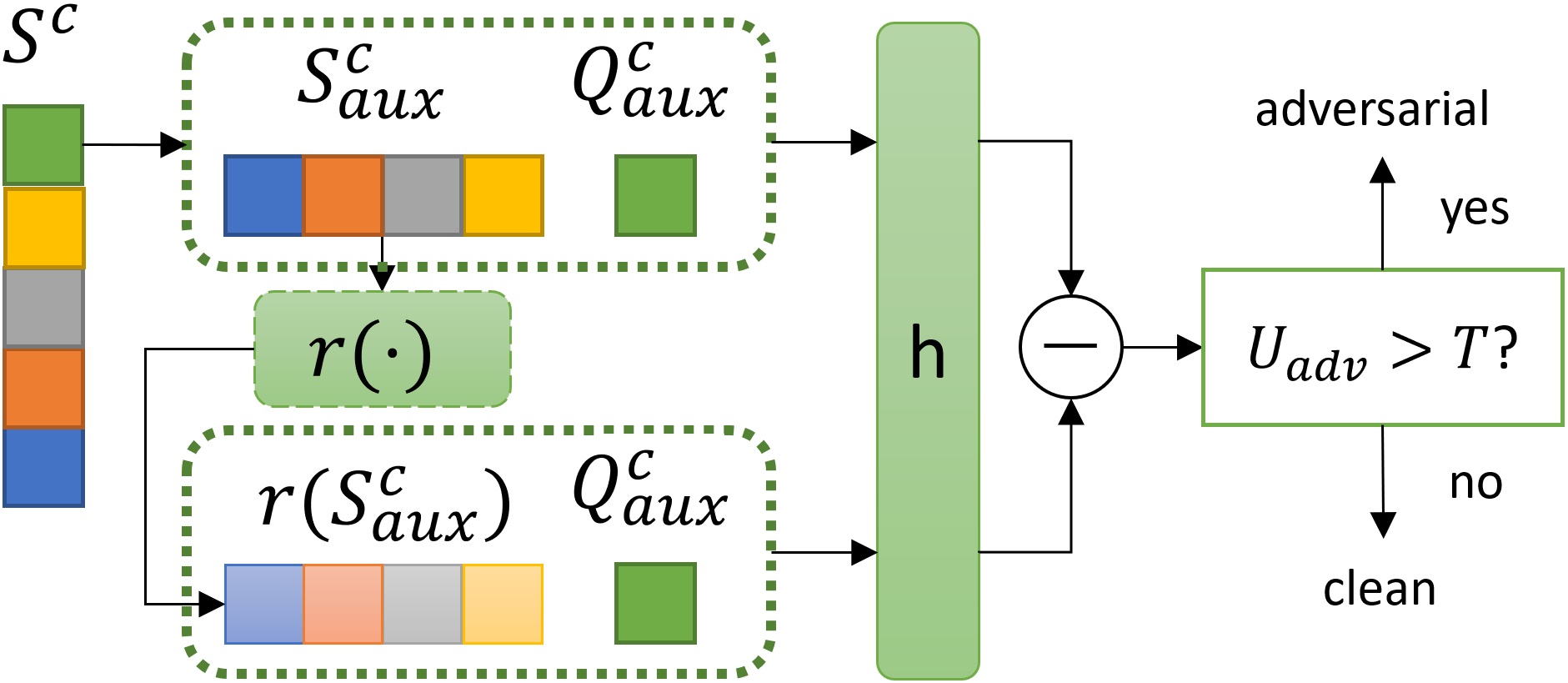}
    \caption{Illustration of our detection mechanism based on self-similarity, by partitioning $S^c$ into two auxiliary sets $S^{c}_{aux}$ and $Q^{c}_{aux}$ and filtering. Best viewed in colour.}
    \label{fig:detectionl1}
\end{figure}

Our detection mechanism flags a support set as adversarial when auxiliary support samples, after filtering, are highly different from the auxiliary query samples, as shown in Figure~\ref{fig:detectionl1}.
Given a support set of class $c$, $S^c$, we split it randomly into two auxiliary sets $S^{c}_{aux}$ and $Q^{c}_{aux}$. 
We filter $S^{c}_{aux}$ using a function $r(\cdot)$ and use the resultant samples as the new auxiliary support set to evaluate $Q^{c}_{aux}$. 
Following which, we obtain the logits of $Q^{c}_{aux}$ both before and after the filtering of the auxiliary support (i.e. using $S^{c}_{aux}$ and $r(S^{c}_{aux})$ respectively) and compute the $\mathcal{L}_1$ norm difference between them. 
The adversarial score $U_{adv}$ is given in \eqref{ssrate} where $h$ is the few-shot classifier
\begin{equation}
\begin{split}
U_{adv} =&\| h(r(S^c_{aux}),Q^c_{aux} ) - h(S^c_{aux},Q^c_{aux} ) \|_1 ,
\label{ssrate}
\end{split}
\end{equation}
and $r$ is any filtering function which maps a support set onto its own space. We observe that we obtain already very high AUROC detection scores when computing $U_{adv}$ without averaging over $n_{shot}$ draws, which we elaborate further later. We flag a support set $S_c$ as adversarial if the adversarial score goes above a certain threshold (i.e.~$U_{adv}>T$). 
Different statistics can be used to compute $U_{adv}$, with Eq.~\eqref{ssrate} being one of many. Our main contribution lies rather in the proposal of using self-similarity of a support set for such detection.

\subsection{Feature-space Preserving Autoencoder (FPA) for Auxiliary Support Set Filtering}
\label{autoencoder}
In light of exploring a DNN-based filtering function, we use an autoencoder (AE) as function $r(\cdot)$ for the detection of adversarial samples in the support set motivated by \cite{folz2020adversarial}. We initially trained a standard autoencoder to reconstruct the clean samples in the image space using the MSE loss. However, the standard autoencoder performed poorly in detecting adversarial supports since it did not learn to preserve the feature space representation of image samples. Therefore, we switched to a feature-space preserving autoencoder
which additionally reconstructs the images in the feature space of the few-shot classifier, contrary to prior work where they fine-tuned their AE on the classification loss \cite{folz2020adversarial}. We argue that using classification loss for fine-tuning is inapplicable in few-shot learning due to having very few labelled samples.
We minimise the following objective function for the feature-space preserving autoencoder: 
\begin{equation}
\mathcal{L_{FPA}} =\frac{1}{N'}\sum^{N'}_{i=1} 0.01 \cdot \ \frac{\| x_i - \hat{x_i} \|_{2}^{2}}{dim(x_i)^{1/2}} \ + \frac{\| f_i - \hat{f_i} \|_{2}^{2}}{dim(f_i)^{1/2}},
\label{aeeqn}
\end{equation}
where $x_i$ and $\hat{x_i}$ are the original and reconstructed image samples, respectively, and $f_i$ and $\hat{f_i}$ are the feature representation of the original and reconstructed image obtained from the few-shot model before any metric module (i.e. features from CNN backbone). The second loss term ensures that the reconstructed image features are similar to those of the original image in the feature space of the few-shot models. We train the feature-space preserving autoencoder by fine-tuning the weights from the standard autoencoder.

\subsection{Median Filtering (FeatS)}
\label{feats}
In our work, we also explored an alternate filtering function. We adopted a feature squeezing (FeatS) filter from \cite{xu2017feature}, where it was used in a conventional classifier. It essentially performs local spatial smoothing of images by having the centre pixel taking the median value among its neighbours within a 2x2 sliding window. As their detection performance was reasonably high using this filter, we decided to use it as an alternative to FPA as an explorative step. However, their approach performs filtering on each individual test sample whereas we use it on the auxiliary support set. 

\subsection{Total Variation Minimisation (TVM)}
\label{tvm}
We have also explored another filtering function, based on the concept of TVM, in our work \cite{guo2018countering}. In essence, the TVM approach performs a reconstruction of randomly selected pixels in the image, selected via a Bernoulli distribution. The reconstruction involves solving an optimisation problem, by minimising the difference between the original and reconstructed images, regularised on the difference between the pixels to the left and above the selected pixel. We made use of the authors' code, as implemented in \cite{guo2018countering}, in our work.

\subsection{Bit Reduction (BitR)}
\label{bitr}
Other than FeatS, we have also adopted a bit reduction (BitR) filter similarly from \cite{xu2017feature}. It essentially reduces the range of values that each pixel in the image can take. For instance, reducing each pixel ($pix$) from an 8-bit precision ($r=8$) to a 4-bit precision ($r=4$) implies reducing the range of values from $pix \in [0,255]$ to $pix \in [0,15]$. Note that this operation was also performed across all colour channels.

\section{Experiments and Results}
\label{fewshotexps}

\subsection{Experimental Settings}

\subsubsection{Datasets}
MiniImagenet (MI) \cite{miniImageNet:vinyals2016matching} and CUB \cite{CUB:wah2011caltech} datasets were used in our experiments. We prepared them following prior benchmark splits \cite{LSTMoptimizer:ravi2016optimization,FeaturewiseTranslayer:tseng2020cross}, with 64/16/20 categories for the train/val/test sets of MI and 100/50/50 categories for the train/val/test sets of CUB. In our attack and detection evaluation, we chose an exemplary set of 10 and 25 classes from the test set for MI and CUB respectively, and we report the average metrics across them. This is purely for computational efficiency. For the RN model, we used image sizes of 224 while using image sizes of 96 for the CAN and PN models across both datasets. 

\subsubsection{Attacks}
In this work, we used two different attack routines, one being PGD while the other being a slight variant of the CW-$\mathcal{L}_2$ attack. This variant uses a normal Stochastic Gradient Descent optimiser instead of Adam as we did not yield good performing adversarial samples with the latter. We still used the objective function defined in
\eqref{chp2:eqncw}
to optimise our CW adversarial samples, while using \eqref{chp2:eqnpgd} to perform a perturbation step less the clipping and sign functions. We name this attack CW-SGD. For our PGD attack, we limit the $L_\infty$ norm of the perturbation ($\epsilon_\infty$) to $12/255$ and a step size of $\eta=0.05$ (see \eqref{chp2:eqnpgd}). 
For our CW-SGD attack, clipping was not used due to the optimisation over $||\delta||_2$ while $\kappa=0.1$ and $\eta=50$. 
We have also evaluated the detection performances on a weaker variant of the above attacks (lower strength attack settings). Namely, for PGD, we limit the $L_\infty$ norm of the perturbation to $6/255$ and $3/255$. For CW-SGD, we explored the settings $\kappa=0$, $\eta=50$ and $\kappa=0$, $\eta=25$.
We would like to stress that optimising for the best set of hyperparameters for generating attacks is not the main focus of our work as we are more interested in obtaining viable adversarial samples. In both settings, we generate 50 sets of adversarial perturbations for each of the 10 and 25 exemplary classes for MI and CUB respectively. We also attack all $n$ support samples for the targeted class $t$. 


\subsubsection{Autoencoder Training Hyperparameters}
We used a ResNet-50 \cite{he2016deep} architecture for the autoencoders\footnote{Autoencoder architecture adapted from GitHub repository https://github.com/Alvinhech/resnet-autoencoder.}. For the MI dataset, we trained the standard autoencoder from scratch with a learning rate of 1e-4. For the CUB dataset, we trained the standard encoder initialised from ImageNet with a learning rate of 1e-4, and the standard decoder from scratch with a learning rate of 1e-3. For fine-tuning of the feature-space preserving autoencoder, we used a learning rate of 1e-4.
We employed a decaying learning rate with a step size of 10 epochs and $\gamma=0.1$. We used the Adam \cite{kingma2014adam} optimiser with a weight decay of 1e-4. In both settings, we used the train split for training and the validation split for selecting our best performing set of autoencoder weights out of 150 epochs of training. It is implemented in PyTorch \cite{paszke2017automatic}.

\subsubsection{Training Few-shot Classifiers Hyperparameters}

We trained the RN and CAN models on the MI and CUB datasets, following the details in \cite{FeaturewiseTranslayer:tseng2020cross} and \cite{CAN:hou2019cross}. The backbone CNNs for RN and CAN models are Resnet-10 and Resnet-12 \cite{he2016deep} respectively. The backbone for the PN model architecture was a simple network with 4-convolutional blocks \cite{PROTO:snell2017prototypical}. All models are trained under the 5-way 5-shot setting.
For the CAN model, we trained with an SGD optimiser for 80 epochs with an initial learning rate of 0.1. We annealed the learning rate to 0.06 at the 60th epoch and 1e-3 at the 70th epoch. For the RN and PN models, we first pre-trained the backbone CNN on the training set as a standard classification task. With the pre-trained CNN, we further trained them with an Adam optimiser and a learning rate of 1e-3 for 400 epochs. We chose the best-performing model on the validation set for the following experiments.

\subsection{Baseline Accuracy of Few-shot Classifiers}

We evaluated our classifiers by taking the average and standard deviation accuracy over 2000 episodes across all models and datasets, reported in Table~\ref{clf:baseline}, to show that we were attacking reasonably performing few-shot classifiers. 

\begin{table}[htb]
\centering
\caption{Baseline classification accuracy of the chosen models on the two datasets, under a 5-way 5-shot setting, computed across 2000 randomly sampled episodes. We report the mean with 95\% confidence intervals for the accuracy.}
\begin{tabular}{lccc}
             \hline
             & RN - 5 shot & CAN - 5 shot & PN - 5 shot \\ \hline
MI & 0.727  $\pm$ 0.0037     & 0.787  $\pm$ 0.0033   &  0.662 $\pm$ 0.0038 \\ \hline
CUB          & 0.842  $\pm$ 0.0032     & 0.890  $\pm$ 0.0026  &  0.724 $\pm$ 0.0038 \\ \hline
\end{tabular}

\label{clf:baseline}
\end{table}

\subsection{Evaluation Metrics Used}

We evaluated the success of our attacks via computing the Attack Success Rate (ASR), measuring the proportion of samples that had adversarial candidates generated from attacks that successfully cause misclassification. 
We only considered samples originating from the targeted class when measuring ASR:
\begin{align}
    ASR=\mathbb{E}_{S^t,Q^t \sim \mathcal{D}}\{P(\mathrm{argmax_y}(h_y(S^t+\delta^t, Q^t)) \neq t)\}.
\end{align}
The remaining $(K-1)$ classes were sampled randomly to make up the support set.

In evaluation of the detection performances, we used the Area Under the Receiver Operating Characteristic (AUROC) metric, since detection problems are binary (whether an adversarial sample is present or not), and true and false positives can be collected at various predefined threshold values.

\subsection{Transferability Attack Results}
\label{transferatt}

\begin{table}[htb]
\centering
\caption{Classification accuracy of the auxiliary query set under various scenarios. ``Clean" indicates that only \textit{clean samples} are present in the support set while ``PGD" and ``CW-SGD" indicates that the respective \textit{adversarial samples} are present in the support set. In the presence of adversarial samples, both the auxiliary support and query set contains adversarial samples. We computed the results across the 10 and 25 exemplary classes for MI and CUB respectively, and 50 sets of adversarial perturbations.}
\begin{tabular}{|c|c|c|c|c|}
\hline
Model & Dataset & Clean & \begin{tabular}[c]{@{}c@{}}PGD\\ ($\epsilon_\infty=12/255$)\end{tabular} & \begin{tabular}[c]{@{}c@{}}CW-SGD\\ ($\kappa=0.1$)\end{tabular} \\ \hline
\multirow{2}{*}{\begin{tabular}[c]{@{}c@{}}RN\\ (5-shot)\end{tabular}} & MI & 0.697 & 0.736 & 0.815 \\
 & CUB & 0.834 & 0.947 & 0.919 \\ \hline
\multirow{2}{*}{\begin{tabular}[c]{@{}c@{}}CAN\\ (5-shot)\end{tabular}} & MI & 0.783 & 1.000 & 1.000 \\
 & CUB & 0.876 & 0.999 & 0.987 \\ \hline
 \multirow{2}{*}{\begin{tabular}[c]{@{}c@{}}PN\\ (5-shot)\end{tabular}} & MI & 0.679  & 0.999  & 0.993   \\
 & CUB &  0.843 & 0.998  & 0.981  \\ \hline
\end{tabular}
\label{tab:selfsim}
\end{table}

\begin{figure*}
    \centering
    \begin{subfigure}[b]{0.48\textwidth}
         \centering
         \includegraphics[trim={2.5cm 15.7cm 3.7cm 2cm},clip,width=\textwidth]{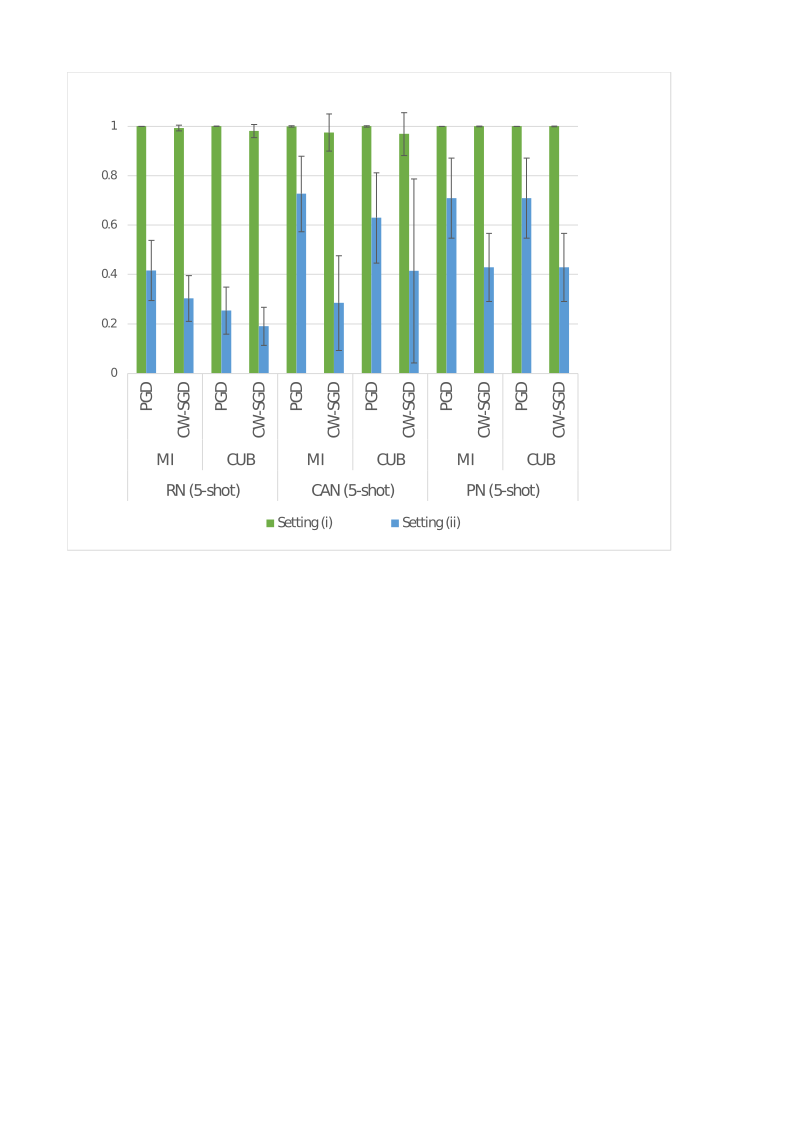}
         \caption{Vanilla attack settings (i.e. $\epsilon_\infty=12/255$ for PGD, $\kappa=0.1$ and $\eta=50$ for CW-SGD).}
         \label{fig:transfervanilla}
     \end{subfigure}
     \hfill
     \begin{subfigure}[b]{0.41\textwidth}
         \centering
         \includegraphics[trim={2.5cm 7.7cm 4.4cm 10cm},clip,width=\textwidth]{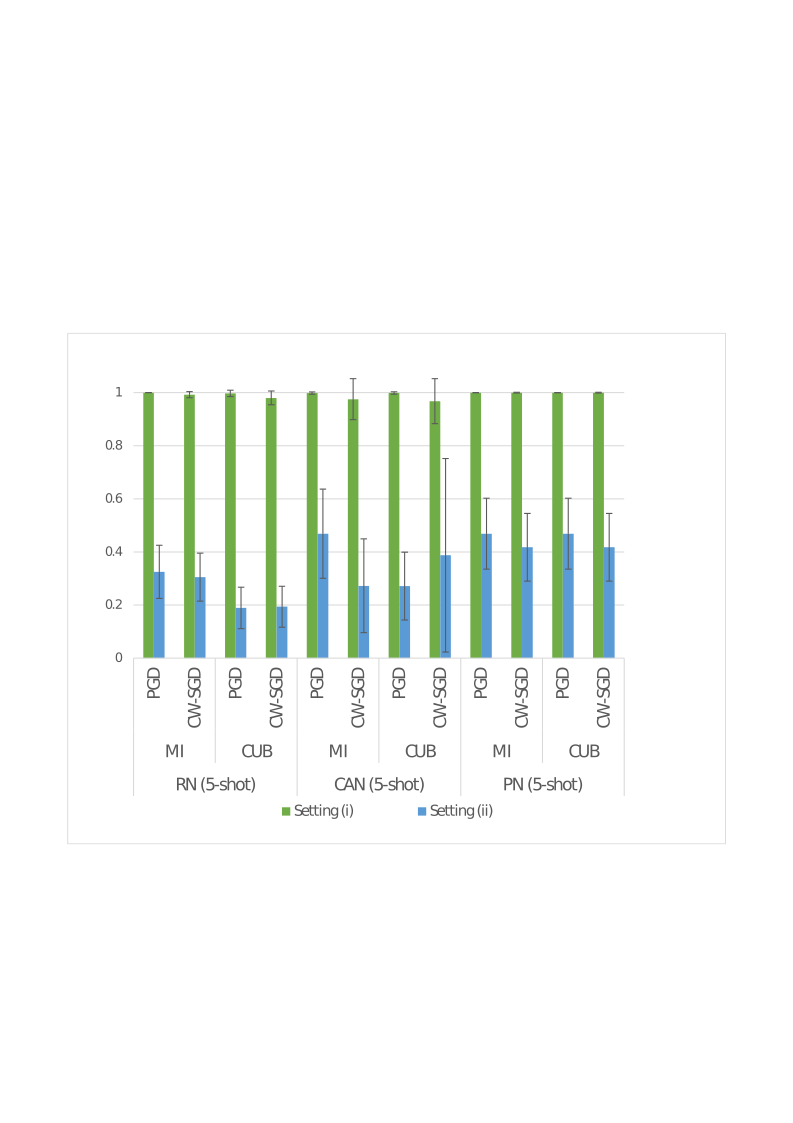}
         \caption{Lower attack settings variant 1 (i.e. $\epsilon_\infty=6/255$ for PGD, $\kappa=0$ and $\eta=50$ for CW-SGD).}
         \label{fig:transferlower}
     \end{subfigure}
     ~
     \begin{subfigure}[b]{0.48\textwidth}
         \centering
         \includegraphics[trim={2.5cm 15.7cm 3.7cm 2cm},clip,width=\textwidth]{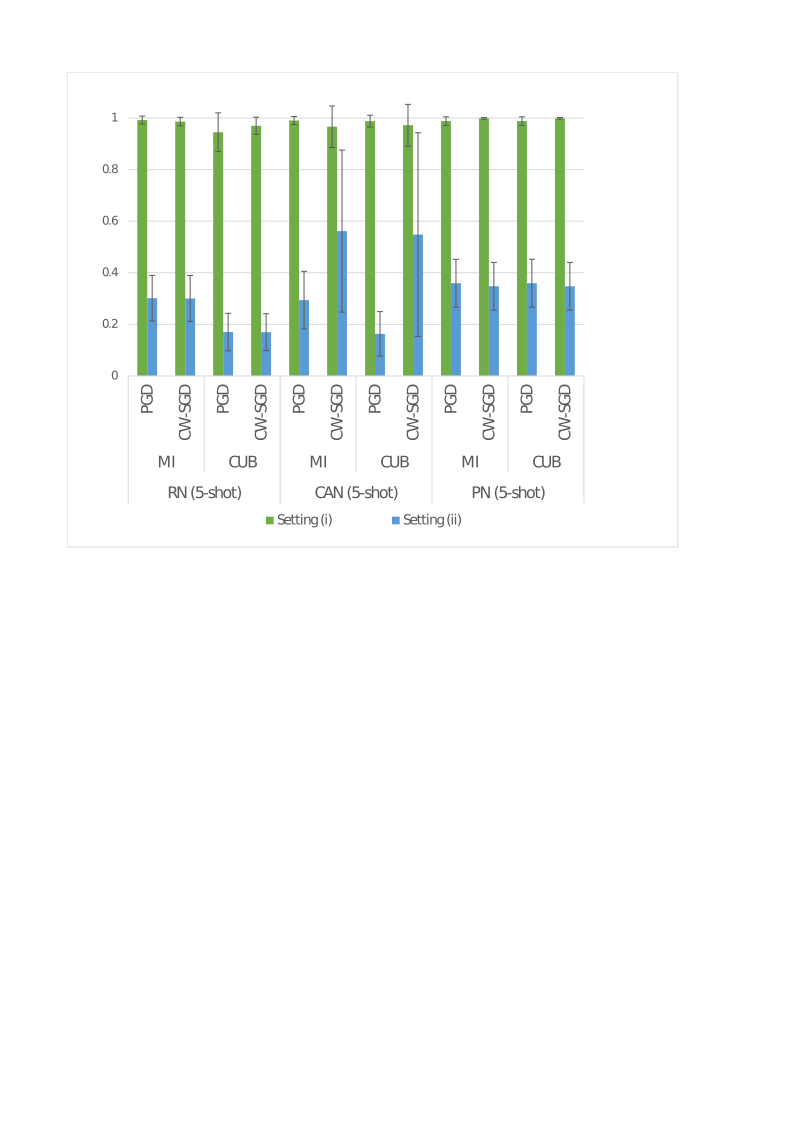}
         \caption{Lower attack settings variant 2 (i.e. $\epsilon_\infty=3/255$ for PGD, $\kappa=0$ and $\eta=25$ for CW-SGD).}
         \label{fig:transferlower2}
     \end{subfigure}
    \caption{Transferability attack results under scenarios (i) Fixed Supports and (ii) New Supports, against our two explored attacks on RN, PN and CAN models across both datasets. Reported ASRs were averaged across the chosen exemplary classes and across the 50 generated sets of adversarial perturbations as bar charts. Standard deviation is represented as the whiskers. ASR metric reported. Attack strength is sorted in the following order: (Figure \ref{fig:transfervanilla}) $>$ (Figure \ref{fig:transferlower}) $>$ (Figure \ref{fig:transferlower2}).}
    \label{fig:transfer}
\end{figure*}

To begin with, the adversarial samples classify each other at a comparable or even better accuracy than clean samples do, and thus cannot be distinguished by looking at accuracy alone.
We show the accuracy of the auxiliary query set when only \textit{clean support set is present} and also when only \textit{adversarial support set is present} in Table~\ref{tab:selfsim}. For the latter, it implies that the adversarial samples are found both in the auxiliary support and query sets. Our results illustrate the self-similar nature of our adversarial support samples, which is not observed for the clean samples.

We conducted transferability experiments to evaluate how well the attacker generalised their generated adversarial perturbation under two unique scenarios: \textit{i) transfer with fixed supports} and \textit{ii) transfer with new supports}. 
Setting (i) assumes that we have the same adversarial support set for class $t$ and we evaluated the ASR over newly drawn query sets. Setting (ii) relaxes this assumption and we instead applied the generated adversarial perturbation, that was stored during the attack phase, on newly drawn support sets for class $t$, similarly evaluating over newly drawn query sets.
Contrary to transferability attacks in conventional setups where a sample is generated on one model and evaluated on another, we performed transferability to new tasks, by drawing randomly sets of non-target classes together with their support sets, and new query sets for the few-shot paradigm.

\begin{figure*}[htb]
    \centering
    \includegraphics[trim={3.cm 1cm 0cm 3.cm},clip,width=1\textwidth]{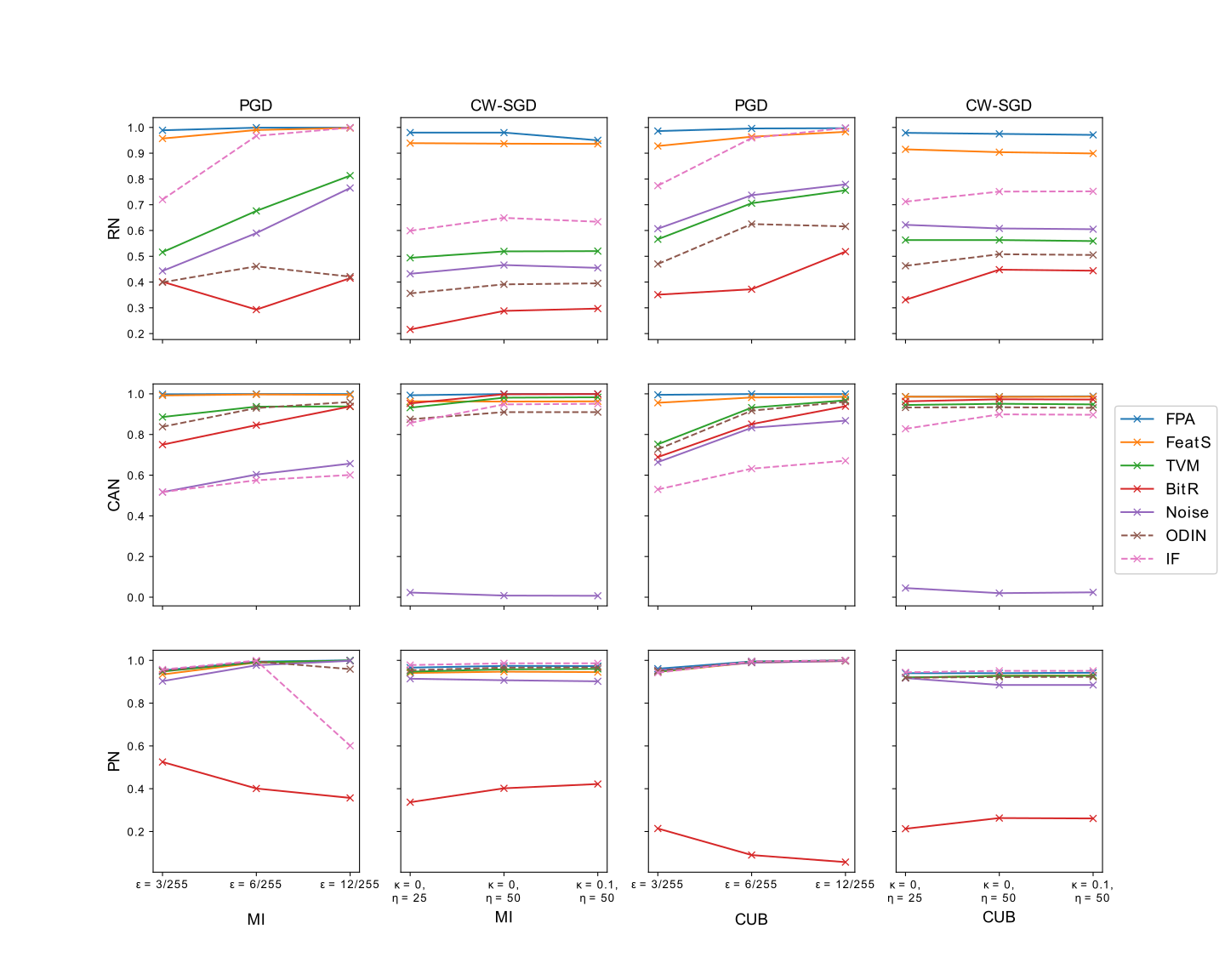}
    \caption{Area Under the Receiver Operator Characteristic (AUROC) scores for the various filter functions (normal distributed noise, median filtering from Feature Squeezing (FeatS), Total Variation Minimisation (TVM), bit reduction @ $r=6$ (BitR), FPA; denoted by solid lines), and our baselines (ODIN and IF; denoted by dashed lines) across our experiment settings, for RN, CAN and PN models. Attack strength increases from left to right within each plot. For PGD attacks, it is determined by the parameter $\epsilon$ while for CW-SGD, it is determined by the parameters $\kappa$ and $\eta$. Higher is better.}
    \label{fig:aucdetection}
\end{figure*}

As illustrated in Figure~\ref{fig:transfer}, the PGD generated adversarial samples showed higher transferability than the CW-SGD attack, across the three models and under all scenarios. The exceptionally high transfer ASR we observed under scenario (i) implies that once the attacker had obtained an adversarial support set targeting a specific class, successful attacks can be carried out on new tasks for which the target class is present. This phenomenon is also present across all the various attack settings, which further reinforces the motivation to investigate defence methods for few-shot classifiers. Under scenario (ii), where the support set of the target class is also randomised, we see lower transfer ASR across the chosen classes. We would like to remind readers that the adversarial samples were optimised explicitly using setting (i) and not for (ii). It is evident that performing attacks under setting (i) (blue bars) is more stable and consistent as compared to setting (ii) (green bars). One can also observe that even against a weaker attacker, the models are similarly as vulnerable under Setting (i), although the degradation in ASR is more apparent in Setting (ii).

\subsection{Detection of Adversarial Supports}
\label{detectadvsupports}

We compared our explored approaches against a simple filtering function for $r(\cdot)$, since prior work of performing detection of adversarial samples in few-shot classifiers does not exist. We experimented with using normal distributed noise as a filter, in which we computed the channel-wise variance for drawing normal distributed noise to be added to the images.
For ODIN, we used $\epsilon_{ODIN}=0.002$ and a temperature $T=100$. We set the hyperparameters as such as recommended by \cite{liang2018enhancing}. 

For IF, we used $4$ isolation trees for the case of CAN while using $1000$ for the case of RN. Our training procedure is as follows: We first split the base data split into two partitions, 90\% for training and 10\% for validation. We also split the attack data that we generated prior into validation and test evaluation using the same ratio, respectively.
Next, we used the training split of the base of MI and CUB to train the respective IF models. We perform hyperparameter fine-tuning on the number of estimators (i.e. number of isolation trees in the forest), by evaluating the validation set (having a mix of attack and clean data). We then select the best performing hyperparameter to evaluate the test set to obtain our results.

Our results in Figure~\ref{fig:aucdetection} shows that FPA exhibits good detection performance across all settings.
Though ``FeatS" also exhibit good detection performances, the FPA approach consistently outperforms it across all settings for RN and CAN, regardless of the attack strength (denoted by the solid blue line chart). 
The ``Noise" approach (denoted by the solid red line chart), however, encountered challenges in detection and is arguably the worse performing among the other filtering functions and outlier detection approaches, when observing the trends for RN and CAN. We see highly varied detection performances across the different settings, which makes this approach highly unreliable.
This result is hardly surprising since such methods require substantial manual fine-tuning of its noise parameters. This is not ideal as newer attacks can be introduced in the future and also, being in a few-shot framework, the optimal noise parameters between different task instances might not be consistent as the data might be different. However, the FPA filter approach exhibits such robustness even in such scenarios as it still achieved favourable AUROC scores. For clean samples, the FPA managed to reconstruct $S_{aux}^c$ such that the logits of $Q_{aux}^c$ before and after filtering remained consistent, even when the FPA did not encounter classes from the novel split during training.
Although our outlier baselines (i.e. ODIN and IF; denoted by the dashed line charts) perform well at times, they can fail at other settings, which also makes them less reliable in performing detection. This phenomenon can also be seen for the BitR filter function, where it performs reasonably well for the CAN model but not for the RN and for the PN models. We note that for BitR, better detection performances could be observed when the detection threshold was flipped (i.e. $U_{adv} > T$ to $U_{adv} < T$). Therefore, we flipped the detection condition for BitR for all models.

\section{Discussion}
\label{discuss}

\subsection{Detection Performance of Attacking Single Sample in Crafting Adversarial Support Sets}

\begin{figure}[htb]
    \centering
    \includegraphics[trim={1.9cm 19.3cm 3.5cm 2.5cm},clip,width=0.47\textwidth]{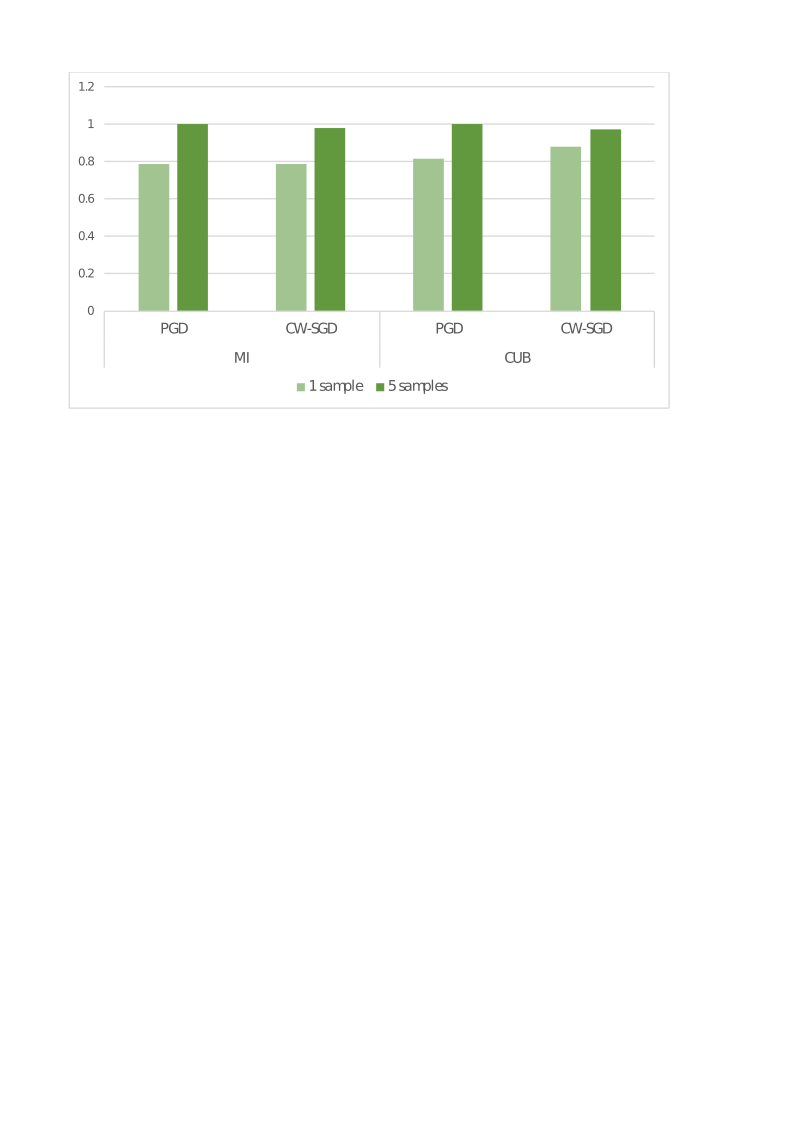}
    \caption{AUROC results comparing between attacking 1 sample and 5 samples in the support set for the CAN, using our FPA to filter the auxiliary support set. We computed the results across the 10 and 25 exemplary classes for MI and CUB respectively, and 50 sets of adversarial perturbations.
    Attack strengths for PGD and CW-SGD were $\epsilon_\infty=12/255$ and $\kappa=0.1$ respectively.
    }
    \label{fig:aucsinglesample}
\end{figure}

We have also explored the effectiveness of our attack detection algorithm when the attacker only attacks a single sample in the support set. We only considered the CAN model in this set of experiments as the behaviour for the case of RN will be similar. We evaluated the AUROC scores when using our FPA filtering function, with Figure \ref{fig:aucsinglesample} illustrating our results. We chose this setting as we wanted to shed some light on the impact of our detection approach, should we adopt another setting (i.e. attacking $N$ samples vs attacking 1 sample). Attackers adopting a scenario between these two settings would yield results which will simply be an interpolation of the two.

\begin{figure}[htb]
    \centering
    \includegraphics[trim={1.9cm 19.2cm 3.6cm 2.2cm},clip,width=0.46\textwidth]{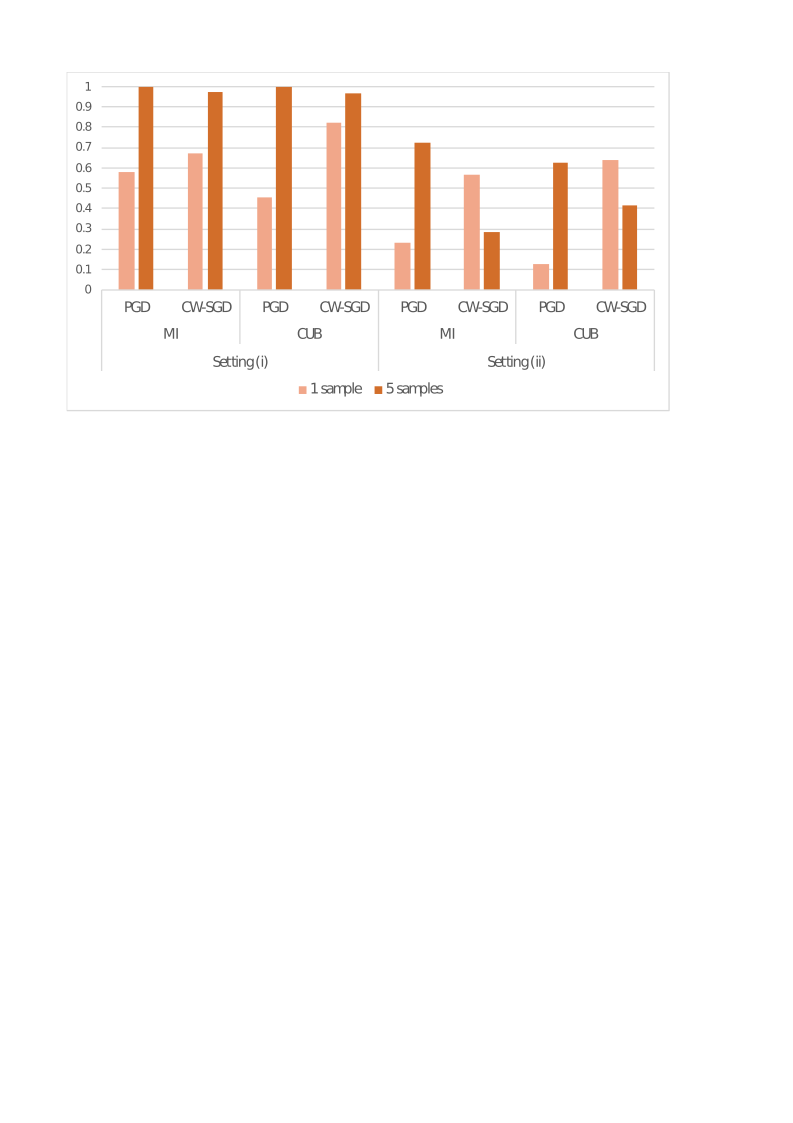}
    \caption{Transferability attack results under settings (i) Fixed Supports and (ii) New Supports, against our two explored attacks on the CAN model across both datasets, comparing between \textit{1 sample} and \textit{5 samples} being attacked in the adversarial support set. Reported ASRs were averaged across the chosen exemplary classes and across the 50 generated sets of adversarial perturbations. Attack strengths for PGD and CW-SGD were $\epsilon_\infty = 12/255$ and $\kappa= 0.1$ respectively. ASR metric reported.}
    \label{fig:asrsinglesamplecompare}
\end{figure}

It is clear that the detection AUROC score suffers as less adversarial samples were found in our support set. This is hardly surprising as there might be instances whereby the adversarial sample was found in the auxiliary query set, rendering the filtering function useless in filtering the adversarial sample. However, even when faced with a single adversarial sample, our algorithm could still detect adversarial supports to a reasonable extent. One could also take the average of multiple random splits of auxiliary support and query sets instead to compute the adversarial score for performing detection to improve the robustness of our detection approach.
Furthermore, attacking a few-shot classifier with only a single sample would not yield favourable attack outcomes as evident in Figure \ref{fig:asrsinglesamplecompare}, where we observe a transferability attack ASR degradation as compared to attacking 5 samples. As such, any detrimental impact attacker inflicts will be lessened as well.

\subsection{Study of Self-Similarity Computation Methods}



\begin{figure}[htb]
    \centering
    \includegraphics[trim={30.3cm 68cm 1cm 1.5cm},clip,width = 0.48\textwidth]{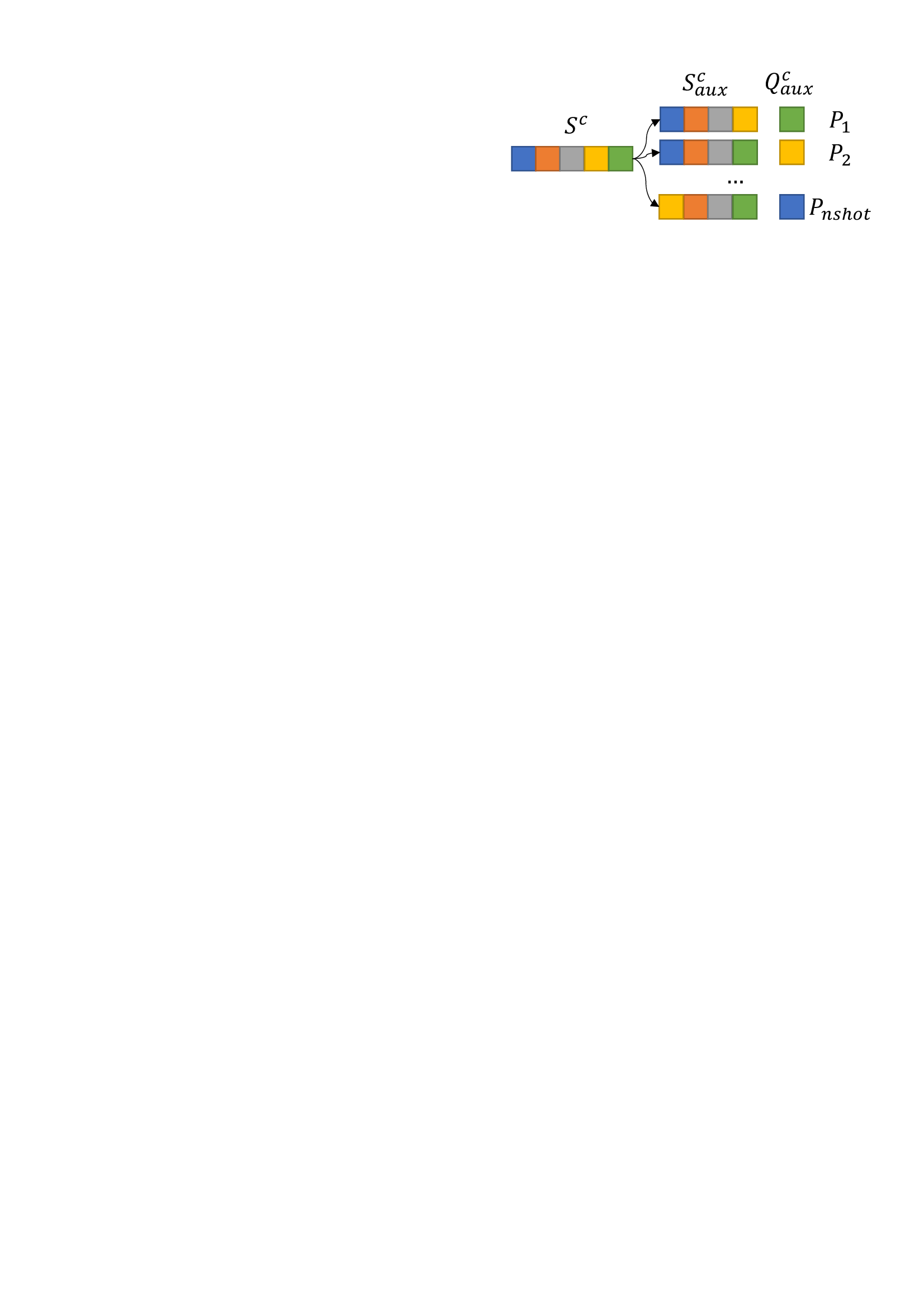}
    \caption{Illustration of how partitioning $S^c$ into two auxiliary sets $S^{c}_{aux}$ and $Q^{c}_{aux}$ is performed. Best viewed in colour.}
    \label{fig:partition}
\end{figure}

In Section~\ref{detectsec}, we described one of the possible detection mechanism based on logits differences. An alternative would be to use hard label predictions. Thus, we investigate the effect of different schemes as a justification for our choice $U_{adv}$. For the case of hard label predictions, we perform the following:
we compute the average accuracy of $Q^{c}_{aux}$, across the different permutated partitions of $S^{c}$. We illustrate how we construct our partitions across the n-shots in Figure \ref{fig:partition}. In essence, each support sample will have a chance to be part of the auxiliary query set.
This results in the statistic $U_{adv}'$:
\begin{equation}
U_{adv}' = \frac{1}{n_{shot}} \sum^{n_{shot}}_{i=1} \mathbbm{1} [\mathrm{argmax}_j(h_j(r(S^{c}_{i, aux}), Q^{c}_{i, aux})) \neq c] ,
\label{ssrate2}
\end{equation}
where $h$ is the few-shot classifier, and $r$ is the filtering function.
Similarly, we flag the support set as adversarial when $U_{adv}'>T$, such that it goes above a certain threshold. 


\begin{table}[htb]
\centering
\caption{Area Under the Receiver Operator Characteristic (AUROC) scores for the two detection mechanisms ($U_{adv}$ and $U_{adv}'$) using our FPA across our experiment settings. Higher is better.}
\begin{tabular}{|c|c|cc|cc|}
\hline
\multirow{2}{*}{Model}                                                  & \multirow{2}{*}{Dataset} & \multicolumn{2}{c|}{\begin{tabular}[c]{@{}c@{}}PGD\\ ($\epsilon_\infty=12/255$)\end{tabular}} & \multicolumn{2}{c|}{\begin{tabular}[c]{@{}c@{}}CW-SGD\\ ($\kappa=0.1$)\end{tabular}} \\ \cline{3-6} 
                                                                        &                          & $U_{adv}$  & $U_{adv}'$    & $U_{adv}$  & $U_{adv}'$     \\ \hline
\multirow{2}{*}{\begin{tabular}[c]{@{}c@{}}RN\\ (5-shot)\end{tabular}}  & MI                       & 0.999	&	0.451		&		0.979 	&	0.723  \\
                                                                        & CUB                      & 	0.997	&	0.326	&	0.974  		& 0.524	 \\ \hline
\multirow{2}{*}{\begin{tabular}[c]{@{}c@{}}CAN\\ (5-shot)\end{tabular}} & MI                       & 0.999	&	0.991	&	0.999	&   0.931	   \\
                                                                        & CUB                      & 0.999		&	0.998	&	0.988	&	0.821  \\ \hline
\multirow{2}{*}{\begin{tabular}[c]{@{}c@{}}PN\\ (5-shot)\end{tabular}}  & MI                       & 0.999 &	0.903 &	0.973 &	0.558	  \\
                                                                        & CUB                      & 0.999 &	0.997 &	0.942 &	0.782\\ \hline
\end{tabular}

\label{tab:aucdetectionDM}
\end{table}

\begin{table}[htb]
\centering
\caption{Area Under the Receiver Operator Characteristic (AUROC) scores for the two detection mechanisms ($U_{adv}$ and $U_{adv}'$) using FeatS across our experiment settings. Higher is better.}
\begin{tabular}{|c|c|cc|cc|}
\hline
\multirow{2}{*}{Model}                                                  & \multirow{2}{*}{Dataset} & \multicolumn{2}{c|}{\begin{tabular}[c]{@{}c@{}}PGD\\ ($\epsilon_\infty=12/255$)\end{tabular}} & \multicolumn{2}{c|}{\begin{tabular}[c]{@{}c@{}}CW-SGD\\ ($\kappa=0.1$)\end{tabular}} \\ \cline{3-6} 
                                                                        &                          & $U_{adv}$                                     & $U_{adv}'$                                    & $U_{adv}$                                & $U_{adv}'$                                \\ \hline
\multirow{2}{*}{\begin{tabular}[c]{@{}c@{}}RN\\ (5-shot)\end{tabular}}  & MI                       & 0.998                                         & 0.585                                         & 0.936                                    & 0.714                                     \\
                                                                        & CUB                      & 0.983                                         & 0.275                                         & 0.899                                    & 0.551                                     \\ \hline
\multirow{2}{*}{\begin{tabular}[c]{@{}c@{}}CAN\\ (5-shot)\end{tabular}} & MI                       & 0.995                                         & 0.304                                         & 0.963                                    & 0.008                                     \\
                                                                        & CUB                      & 0.985                                       & 0.154                                         & 0.986                                  & 0.768                                     \\ \hline
\multirow{2}{*}{\begin{tabular}[c]{@{}c@{}}PN\\ (5-shot)\end{tabular}}  & MI                       & 0.998 &	0.000 &	0.945 &	0.001  \\
                                                                        & CUB                      & 0.997 &	0.000 &	0.929 &	0.004 \\ \hline
\end{tabular}

\label{tab:aucdetectionfeats}
\end{table}

\begin{table}[htb]
\centering
\caption{Area Under the Receiver Operator Characteristic (AUROC) scores for the two detection mechanisms ($U_{adv}$ and $U_{adv}'$) using BitR across our experiment settings. Higher is better.}
\begin{tabular}{|c|c|cc|cc|}
\hline
\multirow{2}{*}{Model}                                                  & \multirow{2}{*}{Dataset} & \multicolumn{2}{c|}{\begin{tabular}[c]{@{}c@{}}PGD\\ ($\epsilon_\infty=12/255$)\end{tabular}} & \multicolumn{2}{c|}{\begin{tabular}[c]{@{}c@{}}CW-SGD\\ ($\kappa=0.1$)\end{tabular}} \\ \cline{3-6} 
                                                                        &                          & $U_{adv}$                                       & $U_{adv}'$                                 & $U_{adv}$                                  & $U_{adv}'$                            \\ \hline
\multirow{2}{*}{\begin{tabular}[c]{@{}c@{}}RN\\ (5-shot)\end{tabular}}  & MI                       & 0.415                                  & 0.002                                      & 0.297                             & 0.006                                 \\
                                                                        & CUB                      & 0.518                                  & 0.004                                      & 0.444                             & 0.001                                 \\ \hline
\multirow{2}{*}{\begin{tabular}[c]{@{}c@{}}CAN\\ (5-shot)\end{tabular}} & MI                       & 0.938                                  & 0.000                                      & 0.999                             & 0.002                                 \\
                                                                        & CUB                      & 0.939                                  & 0.000                                      & 0.972                            & 0.013                                 \\ \hline
\multirow{2}{*}{\begin{tabular}[c]{@{}c@{}}PN\\ (5-shot)\end{tabular}}  & MI                       & 0.357 &	0.286 &	0.422 &	0.240	  \\
                                                                        & CUB                      & 0.057 &	0.153 &	0.261 &	0.114\\ \hline
\end{tabular}

\label{tab:aucdetectionbitr}
\end{table}

\begin{table}[htb]
\centering
\caption{Area Under the Receiver Operator Characteristic (AUROC) scores for the two detection mechanisms ($U_{adv}$ and $U_{adv}'$) using Noise across our experiment settings. Higher is better.}
\begin{tabular}{|c|c|cc|cc|}
\hline
\multirow{2}{*}{Model}                                                  & \multirow{2}{*}{Dataset} & \multicolumn{2}{c|}{\begin{tabular}[c]{@{}c@{}}PGD\\ ($\epsilon_\infty=12/255$)\end{tabular}} & \multicolumn{2}{c|}{\begin{tabular}[c]{@{}c@{}}CW-SGD\\ ($\kappa=0.1$)\end{tabular}} \\ \cline{3-6} 
                                                                        &                          & \multicolumn{1}{c}{$U_{adv}$}                & \multicolumn{1}{c|}{$U_{adv}'$}                & \multicolumn{1}{c}{$U_{adv}$}            & \multicolumn{1}{c|}{$U_{adv}'$}           \\ \hline
\multirow{2}{*}{\begin{tabular}[c]{@{}c@{}}RN\\ (5-shot)\end{tabular}}  & MI                       & 0.765                                        & 0.460                                          & 0.455                                    & 0.562                                     \\
                                                                        & CUB                      & 0.779                                        & 0.200                                          & 0.605                                    & 0.525                                     \\ \hline
\multirow{2}{*}{\begin{tabular}[c]{@{}c@{}}CAN\\ (5-shot)\end{tabular}} & MI                       & 0.657                                        & 0.137                                          & 0.007                                    & 0.000                                     \\
                                                                        & CUB                      & 0.868                                        & 0.437                                          & 0.024                                    & 0.004                                     \\ \hline
\multirow{2}{*}{\begin{tabular}[c]{@{}c@{}}PN\\ (5-shot)\end{tabular}}  & MI                       & 0.998 &	0.551 &	0.902 &	0.558  \\
                                                                        & CUB                      & 0.997 &	0.697 &	0.885 &	0.554 \\ \hline
\end{tabular}

\label{tab:aucdetectionnoise}
\end{table}

\begin{table}[htb]
\centering
\caption{Area Under the Receiver Operator Characteristic (AUROC) scores for the two detection mechanisms ($U_{adv}$ and $U_{adv}'$) using TVM across our experiment settings. Higher is better.}
\begin{tabular}{|c|c|cc|cc|}
\hline
\multirow{2}{*}{Model}                                                  & \multirow{2}{*}{Dataset} & \multicolumn{2}{c|}{\begin{tabular}[c]{@{}c@{}}PGD\\ ($\epsilon_\infty=12/255$)\end{tabular}} & \multicolumn{2}{c|}{\begin{tabular}[c]{@{}c@{}}CW-SGD\\ ($\kappa=0.1$)\end{tabular}} \\ \cline{3-6} 
                                                                        &                          & \multicolumn{1}{c}{$U_{adv}$}                & \multicolumn{1}{c|}{$U_{adv}'$}                & \multicolumn{1}{c}{$U_{adv}$}            & \multicolumn{1}{c|}{$U_{adv}'$}           \\ \hline
\multirow{2}{*}{\begin{tabular}[c]{@{}c@{}}RN\\ (5-shot)\end{tabular}}  & MI                       &  0.813 &	0.891 &	0.520 &	0.764               \\
                                                                        & CUB                      &  0.756 &	0.813 &	0.559 &	0.802                           \\ \hline
\multirow{2}{*}{\begin{tabular}[c]{@{}c@{}}CAN\\ (5-shot)\end{tabular}} & MI                       & 0.938 &	0.814 &	0.983 &	0.861   \\
                                                                        & CUB                      &  0.968 &	0.819 &	0.948 &	0.860                  \\ \hline
\multirow{2}{*}{\begin{tabular}[c]{@{}c@{}}PN\\ (5-shot)\end{tabular}}  & MI                       & 0.999 &	0.799 &	0.960 &	0.737 \\
                                                                        & CUB                      & 0.999 &	0.891 &	0.928 &	0.772 \\ \hline
\end{tabular}

\label{tab:aucdetectiontvm}
\end{table}

Tables~\ref{tab:aucdetectionDM} to \ref{tab:aucdetectiontvm} show our AUROC scores comparing the two detection mechanisms, $U_{adv}$ and $U_{adv}'$. It is evident that using logits scores to calculate differences, as in $U_{adv}$, is more informative than using hard label predictions to match class labels, as $U_{adv}$ outperforms $U_{adv}'$. 
Differences in logits can be pronounced also in cases when the prediction label does not switch. 
We would like to note that when using $U_{adv}'$, for FeatS, BitR, and Noise filters, there is a greater majority of AUROC scores that fall below $0.5$. This indicates that better performance would be achieved when the flagging condition is inverted. However, it will not be experimentally consistent since such inversion should be applied on both $U_{adv}$ and $U_{adv}'$, for any given filter function.

\subsection{Varying Degrees of Regularisation of FPA}
We observe lower AUROC scores for the RN model than the CAN model in Figure~\ref{fig:aucdetection}. As such, we question if this difference can be attributed to the FPA's ability to reconstruct clean samples effectively, as mentioned in the preamble of Section~\ref{fewshotmethod}. Recalling from \eqref{aeeqn}, we define an additional regularisation term to enforce stricter reconstruction requirements to also include class distribution reconstruction. More specifically, we minimise the following objective function: 

\begin{align}
    \begin{split}
    \mathcal{L_{FPA'}}  =\frac{1}{N'}\sum^{N'}_{i=1} 0.01 \cdot \ &\frac{\| x_i - \hat{x_i} \|_{2}^{2}}{dim(x_i)^{1/2}} \ + \frac{\| f_i - \hat{f_i} \|_{2}^{2}}{dim(f_i)^{1/2}} \\
    & \qquad+ \frac{\| z_i - \hat{z_i} \|_{2}^{2}}{dim(z_i)^{1/2}},
    \end{split}
\label{aeeqn2}
\end{align}
where $x_i$ and $\hat{x_i}$ are the original and reconstructed image samples, respectively, and, $f_i$ and $\hat{f_i}$ are the feature representation of the original and reconstructed image obtained from the few-shot model before any metric module, and $z_i$ and $\hat{z_i}$ are the logits of the original and reconstructed image. We refer to this variant as $FPA'$.
Similarly, we train $FPA'$ by fine-tuning the weights from the standard autoencoder.

\begin{table}[htb]
\centering
\caption{AUROC results comparing $FPA$ and $FPA'$ for the RN. We computed the results across the 10 and 25 exemplary classes for MI and CUB respectively, and 50 sets of adversarial perturbations. 
}
\begin{tabular}{|c|cc|cc|}
\hline
\multirow{2}{*}{Dataset} & \multicolumn{2}{c|}{\begin{tabular}[c]{@{}c@{}}PGD\\ ($\epsilon_\infty=12/255$)\end{tabular}} & \multicolumn{2}{c|}{\begin{tabular}[c]{@{}c@{}}CW-SGD\\ ($\kappa=0.1$)\end{tabular}} \\ \cline{2-5} 
                         & $FPA$ & $FPA'$ & $FPA$ & $FPA'$  \\ \hline
MI             &     0.999    &   0.999    &     0.979    &   0.950     \\ \hline
CUB                      &   0.997      &  0.997     &    0.974      &  0.971       \\ \hline
\end{tabular}

\label{tab:scratedirtycompare}
\end{table}

Our results in Table~\ref{tab:scratedirtycompare} shows that surprisingly, imposing a higher degree of regularisation marginally lowers the detection performance of our algorithm rather than improving it. This implies that $FPA$ is already sufficient to induce a large enough divergence in classification behaviours in the presence of an adversarial support set.

\section{Conclusion}
\label{fewshowconclude}

In this work, we made several extensions from our prior work. Firstly, we provide motivation to our conceptually simple approach by analysing the self-similarity of support samples under attack and normal conditions. Secondly, we perform a more in-depth analysis of our detection performance against a wider range of attack strengths and also with an additional few-shot classifier. Thirdly, we provide an analysis of the transferability attack and our detection performance when only a single sample in the support set is targeted. Finally, we study the effects of varying the self-similarity computation method on the detection performance. Through our extended results, we have shown that the FPA approach is still the most effective filtering function (highest AUROC scores) among the explored filter functions, while also being able to outperform simple baseline approaches across our settings. Our algorithm, which uses the concept of self-similarity among samples in the support set and filtering, is thus shown to exhibit some generalizability in essence. Finally, in our single sample attack scenario, we found that although the detection performance drops slightly, the transferability attack results decayed more significantly, which also provides a lower bound of our attack detection performance in essence.

\ifCLASSOPTIONcaptionsoff
  \newpage
\fi



\bibliographystyle{IEEEtran}
\bibliography{biblo}
%



%




\begin{IEEEbiographynophoto}{Yi Xiang Marcus Tan}
graduated with a Ph.D. degree from the Singapore University of Technology and Design (SUTD) in 2021, where he was under the supervision of Alexander Binder and Ngai-Man Cheung during his candidature. His research interest lies in the area of machine learning and how machine learning can be defended against integrity attacks.
\end{IEEEbiographynophoto}

\begin{IEEEbiographynophoto}{Penny Chong}
graduated with a Ph.D. degree from the Singapore University of Technology and Design (SUTD), under the supervision of Alexander Binder and Ngai-Man Cheung. She received a B.Sc. (Hons) in Applied Mathematics with Computing from Universiti Tunku Abdul Rahman (UTAR), Malaysia in 2016. Her research interests include machine learning, explainable AI and its applications.
\end{IEEEbiographynophoto}

\begin{IEEEbiographynophoto}{Jiamei Sun}
graduated with a Ph.D. degree from the Singapore University of Technology and Design in 2021, where she was under the supervision of Alexander Binder during her candidature. Her research interests include machine learning, deep learning and explainable AI.  
\end{IEEEbiographynophoto}

\begin{IEEEbiographynophoto}{Ngai-Man Cheung}
is an associate professor at the Singapore University of Technology and Design. He received his Ph.D. degree in Electrical Engineering from University of Southern California (USC), Los Angeles, CA, in 2008. His research interests include image, video and signal processing, computer vision and AI.
\end{IEEEbiographynophoto}

\begin{IEEEbiographynophoto}{Yuval Elovici}
is the director of the Telekom Innovation Laboratories at Ben-Gurion University of the Negev (BGU), head of BGU Cyber Security Research Center and a professor in the Department of Software and Information Systems Engineering at BGU. He holds a Ph.D. in Information Systems from Tel-Aviv University. His primary research interests are computer and network security, cyber security, web intelligence, information warfare, social network analysis, and machine learning. He is the co-founder of the start-up Morphisec.
\end{IEEEbiographynophoto}

\begin{IEEEbiographynophoto}{Alexander Binder}
is an associate professor at the University of Oslo (UiO). He received a Dr.rer.nat. in Computer Science from Technische  Universit\"{a}t Berlin in 2013. His research interests  include explainable deep learning and medical imaging.
\end{IEEEbiographynophoto}




\end{document}